\documentclass[10pt,twocolumn,letterpaper]{article}

\usepackage{cvpr}              % To produce the CAMERA-READY version
\usepackage{graphicx}
\usepackage{amsmath}
\usepackage{amssymb}
\usepackage{booktabs}
\usepackage{tabulary}
\usepackage[dvipsnames]{xcolor}
\usepackage[pagebackref,breaklinks,colorlinks]{hyperref}

\newcommand{\tablestyle}[2]{\setlength{\tabcolsep}{#1}\renewcommand{\arraystretch}{#2}\centering\footnotesize}

\usepackage[capitalize]{cleveref}
\crefname{section}{Sec.}{Secs.}
\Crefname{section}{Section}{Sections}
\Crefname{table}{Table}{Tables}
\crefname{table}{Tab.}{Tabs.}

\def\cvprPaperID{3256} % *** Enter the CVPR Paper ID here
\def\confName{CVPR}
\def\confYear{2023}

\begin{document}

\title{Dynamic Coarse-to-Fine Learning for Oriented Tiny Object Detection}

\author{Chang Xu$^1$,~
Jian Ding$^2$,~
Jinwang Wang$^1$,~
Wen Yang$^1$\thanks{Corresponding Authors},~
Huai Yu$^1$,~
Lei Yu$^1$\footnotemark[1],~
Gui-Song Xia$^2$ \\
$^1$ School of Electronic Information, Wuhan University\\
$^2$ School of Computer Science, Wuhan University \\
{\tt\small \{xuchangeis,jian.ding,jwwangchn,yangwen,yuhuai,ly.wd,guisong.xia\}@whu.edu.cn}
}
\maketitle

\newcommand{\equspace}{1.0pt}

%%%%%%%%% ABSTRACT
\begin{abstract}
    Detecting arbitrarily oriented tiny objects poses intense challenges to existing detectors, especially for label assignment. Despite the exploration of adaptive label assignment in recent oriented object detectors, the extreme geometry shape and limited feature of oriented tiny objects still induce severe mismatch and imbalance issues. Specifically, the position prior, positive sample feature, and instance are mismatched, and the learning of extreme-shaped objects is biased and unbalanced due to little proper feature supervision.
    To tackle these issues, we propose a dynamic prior along with the coarse-to-fine assigner, dubbed DCFL. For one thing, we model the prior, label assignment, and object representation all in a dynamic manner to alleviate the mismatch issue. For another, we leverage the coarse prior matching and finer posterior constraint to dynamically assign labels, providing appropriate and relatively balanced supervision for diverse instances. 
    Extensive experiments on six datasets show substantial improvements to the baseline. Notably, we obtain the state-of-the-art performance for one-stage detectors on the DOTA-v1.5, DOTA-v2.0, and DIOR-R datasets under single-scale training and testing. Codes are available at \url{https://github.com/Chasel-Tsui/mmrotate-dcfl}.
\end{abstract}

%%%%%%%%% BODY TEXT
\section{Introduction}

The oriented bounding box is a finer representation for object detection since the object's background region is greatly eradicated by introducing the rotation angle~\cite{DOTA_2018_CVPR}. This advantage is pronounced in aerial images, where objects are in arbitrary orientations, resulting in the prosperity of corresponding object detection datasets~\cite{DOTA_2018_CVPR,DOTA2.0_2021_pami,diorr_2022_tgrs,HRSC2016_2016} and customized oriented object detectors~\cite{RoI-Transformer_2019_CVPR,redet_2021_cvpr,SCRDet_2019_ICCV,R3Det_2021_AAAI,s2anet_2021_tgrs}. Nevertheless, one unignorable fact is that there exist numerous tiny objects in aerial images. When oriented objects are tiny-sized, the challenges posed to existing object detectors are quite remarkable. Especially, the extreme geometry characteristics of oriented tiny objects hamper the accurate label assignment.

\begin{figure}[t]
\includegraphics[width=\linewidth]{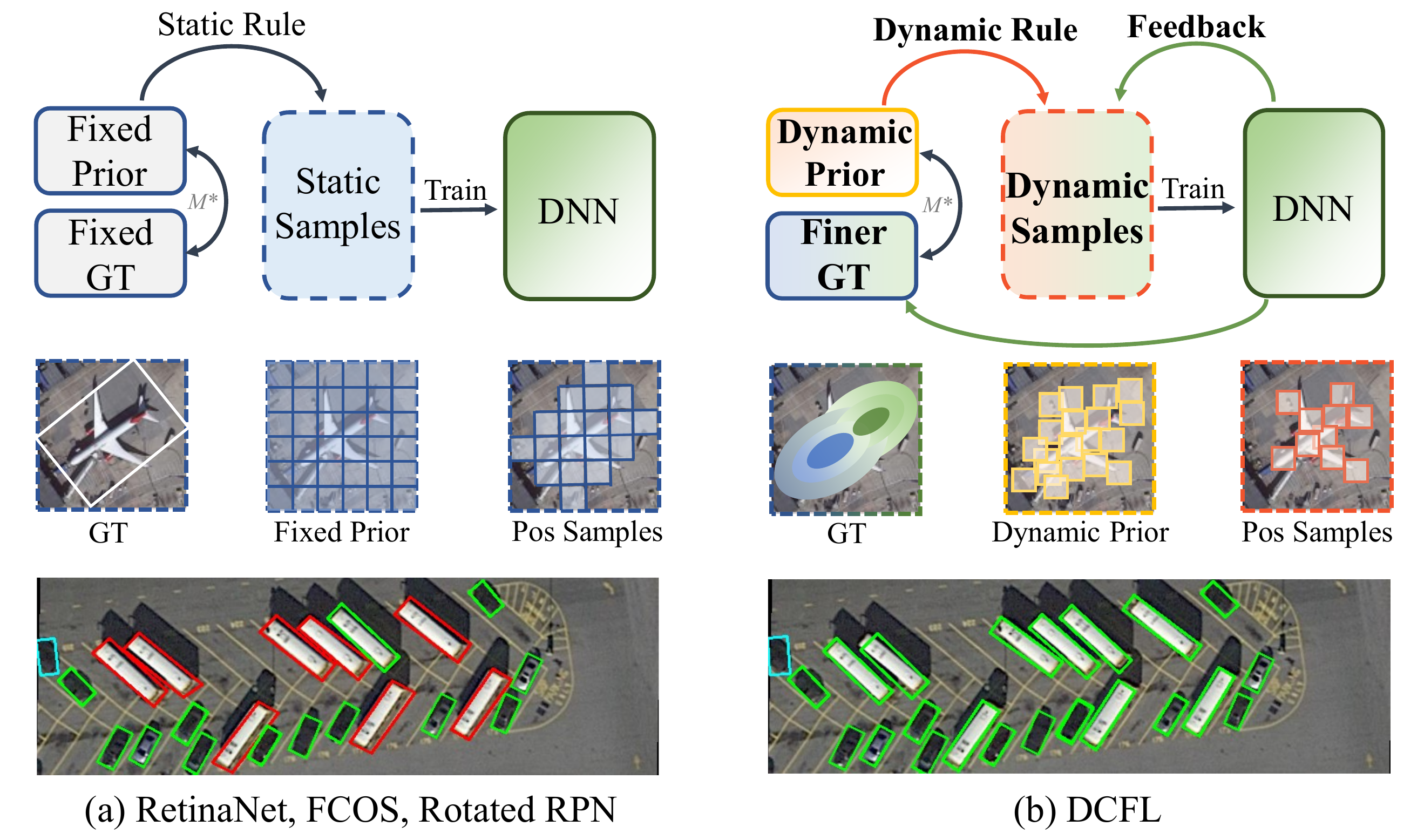}
\caption{Comparisons of different learning paradigms for oriented object detection.~M* means the matching function.~Each box in the $2^{nd}$ row denotes a prior location.~The $3^{rd}$ row are predictions of the RetinaNet and DCFL, where green, blue, and red boxes denote true positive, false positive, and false negative predictions. (a) RetinaNet, FCOS, and Rotated RPN statically assign labels between fixed priors and fixed \textit{gts}. (b) Our proposed DCFL dynamically updates priors and \textit{gts}, and dynamically assigns labels.}
\label{fig:fig1}
\end{figure}

Label assignment is a fundamental and crucial process in object detection~\cite{atss_2020_cvpr}, in which priors (box for anchor-based~\cite{Focal-Loss_2017_ICCV} and point for anchor-free detectors~\cite{FCOS_2019_ICCV}) need to be assigned with appropriate labels to supervise the network training. 
In fact, there have been some works that lay a foundation for the effective label assignment of oriented objects, as shown in Fig.~\ref{fig:fig1}.
Early works additionally preset anchors of different angles (\textit{e.g.}~Rotated RPN~\cite{rotatedrpn_2018_tmm}) or refine high-quality anchors (\textit{e.g.}~$\rm S^{2}A$-Net~\cite{s2anet_2021_tgrs}) based on the generic object detector, then a static rule (\textit{e.g.} MaxIoU strategy~\cite{Faster-R-CNN_2015_NIPS}) is used to separate positive and negative (\textit{pos/neg}) training samples. The derived prior boxes can thus cover more ground truth (\textit{gt}) boxes and a considerable accuracy improvement can be expected. However, the static assignment cannot adaptively divide \textit{pos/neg} samples according to the \textit{gt's} shape and filter out low-quality samples, usually leading to sub-optimal performance.

Recently, the exploration of adaptive label assignment~\cite{atss_2020_cvpr} brings new insight to the community. For oriented object detection, DAL~\cite{dal_2021_aaai} defines a prediction-aware matching degree and utilizes it to reweight anchors, achieving dynamic sample learning.  
Besides, several studies~\cite{sasm_2022_aaai,gghl_2022_tip,orientedrep_2022_cvpr} incorporate the shape information into detectors and propose shape-aware sampling and measurement. 

\begin{figure}[t]
\includegraphics[width=\linewidth]{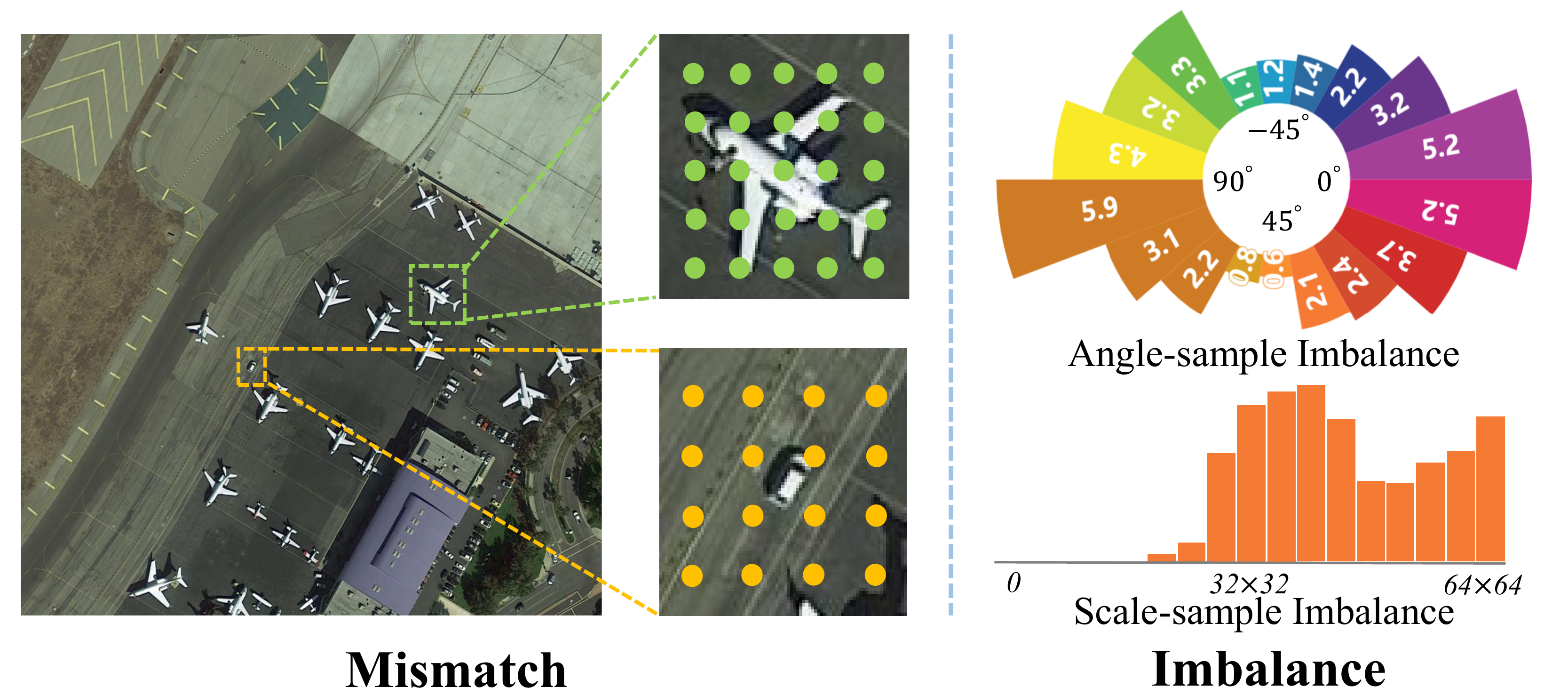}
\caption{The mismatch and imbalance issues. Each point in the left image denotes a prior location. The number in the pie-shaped bar chart denotes the mean number of positive samples assigned to each instance in a specific angle range.}
\label{fig:issues}
\end{figure}

Despite the progress, the arbitrary orientation and extreme size of oriented tiny objects still pose a dilemma to the detector. As shown in Fig.~\ref{fig:issues}, the \textbf{mismatch} and \textbf{imbalance} issues are particularly pronounced. 
For one thing, there is a mutual mismatch issue between the \textit{position prior}, \textit{feature}, and \textit{instance}.
Although some adaptive label assignment schemes may explore a better \textit{pos/neg} division of the prior boxes or points, the sampled feature location behind the prior is still fixed and the derived prior is still static and uniformly located, most priors deviate from the tiny object's main body. The prior and feature themselves cannot well-match the extreme shapes of oriented tiny objects, no matter how we divide \textit{pos/neg} samples.~For another, the existing detectors tend to introduce bias and imbalance for oriented and tiny objects. More precisely, for anchor-based detectors, \textit{gt} with shapes different from anchor boxes will yield low IoU\cite{dal_2021_aaai,dotd_2021_cvprw}, leading to the lack of positive samples. In Fig.~\ref{fig:issues}, we calculate the mean number of positive samples assigned to different \textit{gts} with the RetinaNet and observe that there is an extreme lack of positive samples for \textit{gts} with angles and scales far from predefined anchors.
For anchor-free detectors, the static prior and its fixed stride limit the upper number of high-quality positive samples. Tiny objects only cover a limited number of feature points, and most of these points are away from the object's main body.  

This motivates us to design a more dynamic and balanced learning pipeline for oriented tiny object detection. As shown in Fig.~\ref{fig:fig1}, we alleviate the mismatch issue by reformulating the prior, label assignment, and \textit{gt} representation all in a dynamic manner, which can be updated by the Deep Neural Network (DNN). Simultaneously, we dynamically and progressively assign labels in a coarse-to-fine manner to seek balanced supervision for various instances. 

Specifically, we introduce a dynamic Prior Capturing Block (PCB) to learn the prior, which adaptively adjusts the prior location while retaining the physical meaning of prior~\cite{anchordetr_2022_aaai}. The PCB is inspired by the paradigm of learnable proposals in the DETR~\cite{detr_2020_eccv} and Sparse R-CNN~\cite{sparsercnn_2021_cvpr} which naturally avoids the mismatch issue between the predefined prior and feature. Compared to this paradigm, we introduce its flexibility for prior updates while keeping the fast-convergence ability of dense detectors~\cite{anchordetr_2022_aaai,dabdetr_2021_iclr}.
Based on the dynamic prior, we then select Cross-FPN-layer Coarse Positive Sample (CPS) candidates for further label assignment, and the CPS is realized by the Generalized Jensen-Shannon Divergence~\cite{gjsd_2020_entropy} (GJSD) between the \textit{gt} and the dynamic prior. The GJSD is able to enlarge the CPS to the object's nearby spatial locations and adjacent FPN layers, ensuring more candidates for extreme-shaped objects.
After obtaining the CPS, we re-rank these candidates with predictions (posterior) and represent the \textit{gt} with a finer Dynamic Gaussian Mixture Model (DGMM), filtering out low-quality samples. 
All designs are incorporated into an end-to-end one-stage detector without additional branches. 

In short, our contributions are listed as follows: (1) We identify that there exist severe mismatch and imbalance issues in the current learning pipeline for oriented tiny object detection. (2) We design a Dynamic Coarse-to-Fine Learning (DCFL) scheme for oriented tiny object detection, which is the first to model the prior, label assignment, and \textit{gt} representation all in a dynamic manner. In the DCFL, we propose to use the GJSD to construct Coarse Positive Samples (CPS) and represent objects with a finer Dynamic Gaussian Mixture Model (DGMM), obtaining coarse-to-fine label assignment. 
(3) Extensive experiments on six datasets show promising results.

\section{Related Work}
\subsection{Oriented Object Detection}
\textbf{Prior for Oriented Objects.}~Anchor, as a classic design in generic object detectors (\textit{e.g.}~Faster R-CNN~\cite{Faster-R-CNN_2015_NIPS}, RetinaNet~\cite{Focal-Loss_2017_ICCV}), has facilitated object detection for a long time. Similarly, oriented object detection also benefits from the anchor design. Initially, rotated RPN~\cite{rotatedrpn_2018_tmm} extends the RPN to the field of oriented object detection by tiling 54 anchors each location with preset angles and scales. Indeed, enumerating potential \textit{gt} shapes can notably improve the recall, apart from the sacks of additional computational cost. RoI Transformer~\cite{RoI-Transformer_2019_CVPR} utilizes horizontal anchors and transforms the RPN-generated horizontal proposals to oriented proposals, reducing the number of rotated anchors. To save computation, the Oriented R-CNN~\cite{orientedrcnn_2021_iccv} introduces~an~oriented RPN that directly predicts oriented proposals based on horizontal anchors. 
Recently, one-stage oriented object detectors gradually emerged, including anchor-based detectors~\cite{R3Det_2021_AAAI,s2anet_2021_tgrs} with box prior and anchor-free detectors~\cite{fcosr_2021_arxiv,orientedrep_2022_cvpr} with point prior. Most of them retain the fixed prior design, except for the $\rm{S^{2}A}$-Net~\cite{s2anet_2021_tgrs} which proposes to generate high-quality anchors.   

\textbf{Label Assignment.} ATSS~\cite{atss_2020_cvpr} reveals that label assignment plays a pivotal role in the detectors' performance~\cite{paassignment_2020_eccv,ota_2021_cvpr,iqdet_2021_cvpr}. In the field of oriented object detection, DAL~\cite{dal_2021_aaai} observes inconsistency between the input prior IoU and the output predicted IoU, then defines a matching degree as the soft label that dynamically reweights anchors. Recently, SASM~\cite{sasm_2022_aaai} introduces a shape-adaptive sample selection and measurement strategy to improve detection performance.  
Similarly, GGHL~\cite{gghl_2022_tip} proposes to fit the main body of the instance by a single 2-D Gaussian heatmap, then it divides and reweights samples in a dynamic manner. In addition, Oriented Reppoints~\cite{orientedrep_2022_cvpr} improves the RepPoints~\cite{RepPoints_2019_ICCV} by assessing the quality of points for more effective label assignment.

\subsection{Tiny Object Detection}

\textbf{Multi-scale Learning.} Basically, one can use a multi-resolution image pyramid to obtain multi-scale learning. However, the vanilla image pyramid will bring much computation cost. 
Thus, some works~\cite{SSD_2016_ECCV,FPN_2017_CVPR,M2Det_2019_AAAI,PANet_2018_CVPR,Efficientdet_2020_CVPR,DetectoRS_2020_CVPR} reduce computation with the efficient Feature Pyramid Network (FPN). 
Unlike the FPN, TridentNet~\cite{Trident-Net_2019_ICCV} introduces multi-branch detection heads of various receptive fields for multi-scale prediction. Moreover, one can normalize the scale of objects for scale-invariant object detection, for example, SNIP~\cite{SNIP_2018_CVPR} and SNIPER~\cite{SNIPER_2018_NIPS} resize images and train objects within a certain scale range.

\textbf{Label Assignment.} Tiny objects usually have low IoU with anchors or cover a limited number of feature points, thus suffering from the lack of positive samples. ATSS~\cite{atss_2020_cvpr} slightly reconciles the number of positive samples for objects of different scales. NWD~\cite{aitodv2_2022_isprs} designs a new metric to replace IoU, which can sample more positive samples for tiny objects. Recently, the RFLA~\cite{rfla_2022_eccv} utilizes outliers to detect tiny objects for scale-balanced learning. 

\textbf{Context Information.} Tiny object lacks discriminative features, but objects are closely related to the surrounding context. Therefore, we can leverage the context information to enhance small object detection. Muti-Region CNN (MRCNN)~\cite{mrcnn_2015_cvpr} and Inside-Outside Network (ION)~\cite{Inside_Outside_Net_2016_CVPR} are two representative works that exploit local and global context information. Recently, the Relation Network~\cite{relationnet_2018_cvpr} and transformer-based detectors~\cite{detr_2020_eccv,anchordetr_2022_aaai,deformabledetr_2021_iclr} reason about the association between instances via the attention mechanism.

\textbf{Feature Enhancement.} The feature representation of small objects can be enhanced by super-resolution or GAN. PGAN~\cite{PGAN_2017_CVPR} first applies GAN to small object detection. Besides, Bai~\textit{et al.}~\cite{SOD-MTGAN_2018_ECCV}~introduce the MT-GAN which trains an image-level super-resolution model to improve the RoI features of small objects. In addition, there are some other methods based on super-resolution including ~\cite{Better_to_Follow_2019_ICCV,auxiliarygan_2021_rs,residualsuperres_2021_rs,edgegan_2020_rs}.

By contrast, our method simultaneously handles the prior mismatch and unbalanced learning via dynamically modeling the prior, label assignment, and \textit{gt} representation. Meanwhile, unlike the two-stage RoI-Transformer~\cite{RoI-Transformer_2019_CVPR} or one-stage $\rm{S^{2}A}$-Net~\cite{s2anet_2021_tgrs}, we embed the dynamic prior inside the end-to-end one-stage detector without introducing any auxiliary branch. 

\section{Method}
\textbf{Overview.} Given a set of dense prior $P \in \mathbb{R}^{W \times H \times C}$ ($W \times H$ is the feature map size, $C$ is the shape information number, each feature point has one prior for simplicity), object detectors remap the set $P$ into final detection results $D$ through the Deep Neural Network (DNN), which can be simplified as:
\begin{equation}
    D = \mathrm{DNN}_{h}(P),
    \label{remap}
\end{equation}
where $\mathrm{DNN}_{h}$ denotes the detection head. Detection results $D$ contain two parts: classification scores $D_{cls} \in \mathbb{R}^{W \times H \times A}$ ($A$ is the class number) and box locations $D_{reg} \in \mathbb{R}^{W \times H \times B}$ ($B$ is the box parameter number).

To train the $\mathrm{DNN}_{h}$, we need to find a proper matching between the prior set $P$ and the \textit{gt} set $GT$, and assign \textit{pos/neg} labels to $P$ to supervise the network learning. For static assigners (\textit{e.g.}~RetinaNet~\cite{Focal-Loss_2017_ICCV}), the set of \textit{pos} labels $G$ can be obtained via hand-crafted matching function $\mathcal{M}_s$: 
\begin{equation}
    G = \mathcal{M}_{s}(P, GT).
    \label{static_mapping}
\end{equation}

For dynamic assigners~\cite{paassignment_2020_eccv,ota_2021_cvpr,dal_2021_aaai}, they tend to simultaneously leverage the prior information $P$ and posterior information (predictions) $D$, and then apply a prediction-aware mapping $\mathcal{M}_d$ to get the set $G$:
\begin{equation}
     G = \mathcal{M}_{d}(P, D, GT).
    \label{dynamic_mapping}   
\end{equation}

After the \textit{pos/neg} label separation, the loss function can be summarized into two parts:
\begin{equation}
    \mathcal{L}= \sum_{i=1}^{N_{pos}} \mathcal{L}_{pos}(D_{i}, G_{i}) +  \sum_{j=1}^{N_{neg}} \mathcal{L}_{neg}(D_{j}, y_j),
    \label{basic_loss}
\end{equation}
where $N_{pos}$, $N_{neg}$ are the number of positive and negative samples respectively, $y_j$ denotes the negative label.

While in this work, we model the prior, label assignment, and \textit{gt} representation all in a dynamic manner to alleviate the mismatch issue. To begin with, the dynamic prior is reformulated to (~$\Tilde{}$ denotes the dynamic item):
\begin{equation}
    \Tilde{D} = \mathrm{DNN}_{h}(\underbrace{\mathrm{DNN}_{p}(P)}_{\text{Dynamic Prior}~\Tilde{P}}), 
\end{equation}
$\mathrm{DNN}_{p}$ is a learnable block incorporated within the detection head to update the prior. Then, the matching function is reformulated to a coarse-to-fine paradigm:
\begin{equation}
    \Tilde{G} = \mathcal{M}_{d}(\mathcal{M}_s(\Tilde{P}, GT), \Tilde{GT}),
    \label{c2f_mapping}
\end{equation}
the $\Tilde{GT}$ is a finer representation of an object with the Dynamic Gaussian Mixture Model (DGMM). In a nutshell, our final loss is modeled as:
\begin{equation}
    \setlength{\abovedisplayskip}{\equspace}
	\setlength{\belowdisplayskip}{\equspace}
    \mathcal{L}= \sum_{i=1}^{\Tilde{N}_{pos}} \mathcal{L}_{pos}(\Tilde{D}_{i}, \Tilde{G}_{i}) +  \sum_{j=1}^{\Tilde{N}_{neg}} \mathcal{L}_{neg}(\Tilde{D}_{j}, y_j).
    \label{new_loss}
\end{equation}
 %specifically, $\mathcal{L}_{cls}$ is focal loss and $\mathcal{L}_{reg}$ is IoU loss.

\begin{figure*}[t]
\centering
\includegraphics[width=\linewidth]{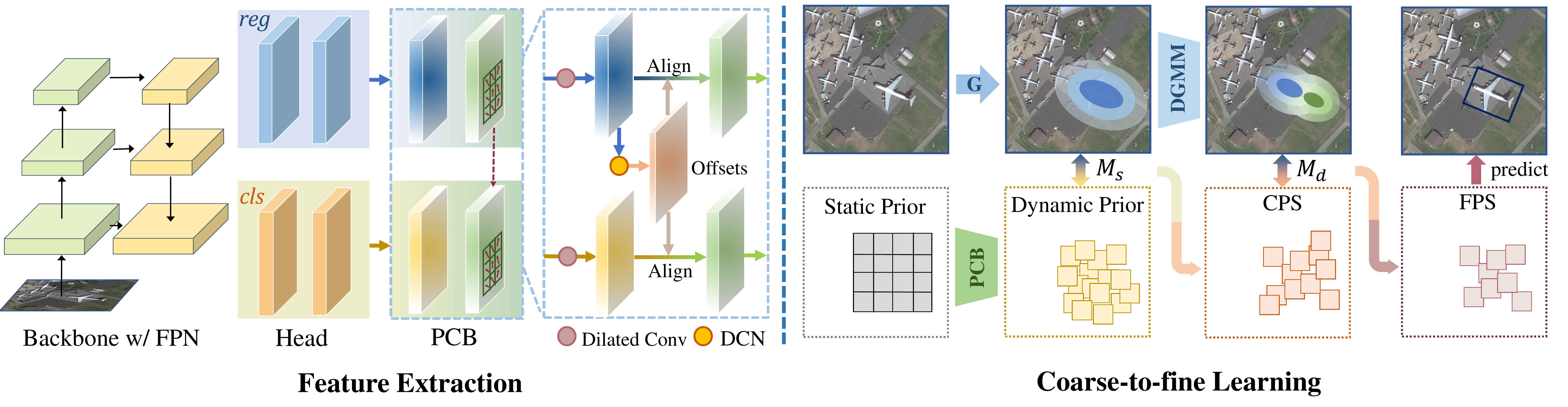}
\caption{The process of feature extraction and dynamic coarse-to-fine learning. PCB denotes the prior capturing block.}
\label{fig:pcb}
\end{figure*}

\subsection{Dynamic Prior}
Inspired by the purely learnable paradigm of proposal updation in the DETR~\cite{detr_2020_eccv} and Sparse R-CNN~\cite{sparsercnn_2021_cvpr}, we propose to introduce more flexibility into the prior to mitigate the mismatch issue. Moreover, we retain the physical meaning of prior where each individual prior stands for a feature point, inheriting the fast convergence ability of dense detectors. The structure of the proposed Prior Capturing Block (PCB) is shown in Fig.~\ref{fig:pcb}, in which a dilated convolution is deployed to take the surrounding information into account, and then the Deformable Convolution Network (DCN)~\cite{DCN_2017_CVPR} is leveraged to capture the dynamic prior. Besides, we utilize the learned offsets from the regression branch to guide the feature extraction of the classification branch, leading to better alignment between the two tasks.

The dynamic prior capturing process is as follows. First of all, we initialize each prior location $\mathbf{p}(x, y)$ by each feature point's spatial location $\mathbf{s}$ (which is remapped to the image). In each iteration, we forward the network to capture the offset sets of each prior location $\Delta\mathbf{o}$. Hence, the prior spatial location can be updated by:
\begin{equation}
    \setlength{\abovedisplayskip}{\equspace}
	\setlength{\belowdisplayskip}{\equspace}
    %\Tilde{\mathbf{s}} = \mathbf{s} + \overbrace{st \sum_{i=1}^{n} \Delta \mathbf{o}_i/ 2n}^{\Delta \mathbf{s}}
    \Tilde{\mathbf{s}} = \mathbf{s} +st \sum_{i=1}^{n} \Delta \mathbf{o}_i/ 2n,
    \label{update_prior}
\end{equation}
where $st$ is the feature map's stride, $n$ is the number of offsets. Finally, we utilize the 2-D Gaussian distribution $\mathcal{N}_p(\boldsymbol{\mu}_p, \boldsymbol{\Sigma}_p)$ which is demonstrated conducive to small objects~\cite{rfla_2022_eccv,gwd_2021_icml} and oriented objects~\cite{gwd_2021_icml,kld_2021_nips} to fit the prior spatial location. Concretely, the dynamic $\Tilde{\mathbf{s}}$ serves as the Gaussian's mean vector $\boldsymbol{\mu}_p$. We preset one prior which is square-shaped $(w, h, \theta)$ as that in RetinaNet~\cite{Focal-Loss_2017_ICCV} on~each feature point, then compute the co-variance matrix $\boldsymbol{\Sigma}_p$ by~\cite{gaussian3d_2022_pami}:

\begin{equation}\small
    \setlength{\abovedisplayskip}{\equspace}
	\setlength{\belowdisplayskip}{\equspace}
    \mathbf{\Sigma}_p=\begin{bmatrix}
        \cos{\theta} & -\sin{\theta} \\
        \sin{\theta} & \cos{\theta}
    \end{bmatrix}\begin{bmatrix}
    \frac{w^2}{4} & 0 \\ 
    0 & \frac{h^2}{4}
    \end{bmatrix}\begin{bmatrix}
        \cos{\theta} & \sin{\theta} \\
        -\sin{\theta} & \cos{\theta}
    \end{bmatrix}.
    \label{gaussian_model}
\end{equation}

\subsection{Coarse Prior Matching}

Given a set of prior, one basic assignment rule is to specify a range of candidate true prediction samples for a specific \textit{gt}. Some adaptive strategies restrict the candidates of a given \textit{gt} inside a single FPN layer~\cite{atss_2020_cvpr,gghl_2022_tip,ota_2021_cvpr}, while some works release all layers as candidates~\cite{objectbox_2022_eccv,fsaf_2019_cvpr}. However, for oriented tiny objects, the former strict heuristic rule may lead to a sub-optimal layer selection and the latter loose one will induce the slow convergence issue~\cite{dabdetr_2021_iclr}.

Hence, we propose Cross-FPN-layer Coarse Positive Sample (CPS) candidates, which narrows down the sample range compared to the all-FPN-layer manner while discarding the single-layer heuristic. In the CPS, we slightly expand the range of candidates to the \textit{gt}'s nearby spatial location and adjacent FPN layers, which warrants relatively diverse and sufficient candidates compared to the single-layer heuristic and alleviates the quantity imbalance issue.

Specifically, the similarity measurement in constructing the CPS is realized with the Jensen-Shannon Divergence (JSD)~\cite{JS_divergence_TIT_2003}, which inherits the scale invariance property of the Kullback–Leibler Divergence (KLD)~\cite{kld_2021_nips} and can measure the \textit{gt}'s similarity with nearby non-overlapping priors~\cite{kld_2021_nips,rfla_2022_eccv}. Moreover, it conquers KLD's drawback of asymmetry. However, the closed-form solution of the JSD between Gaussian distributions is unavailable~\cite{gjsd_2020_entropy}, thus, we utilize the Generalized Jensen-Shannon Divergence (GJSD)~\cite{gjsd_2020_entropy} which yields a closed-form solution, as the substitute.

For example, the GJSD between two Gaussian distributions $\mathcal{N}_{p}(\boldsymbol{\mu}_{p}, \boldsymbol{\Sigma}_{p})$ and $\mathcal{N}_{g}(\boldsymbol{\mu}_{g}, \boldsymbol{\Sigma}_{g})$ is defined by:
{
\begin{equation}
\begin{aligned}
        \mathrm{GJSD}(\mathcal{N}_p, \mathcal{N}_g) &= (1-\alpha) \mathrm{KL}(\mathcal{N}_{\alpha}, \mathcal{N}_{p}) + \alpha \mathrm{KL}(\mathcal{N}_{\alpha}, \mathcal{N}_{g}), \\
                      %&= \frac{1}{2}\left((1-\alpha) \mu_1^{\top} \Sigma_1^{-1} \mu_1+\alpha \mu_2^{\top} \Sigma_2^{-1} \mu_2-\mu_\alpha^{\top} \Sigma_\alpha^{-1} \mu_\alpha+\log \frac{\left|\Sigma_1\right|^{1-\alpha}\left|\Sigma_2\right|^\alpha}{\left|\Sigma_\alpha\right|}\right)
\end{aligned}
\end{equation}
}where $\mathrm{KL}$ denotes the KLD, and $\mathcal{N}_{\alpha}(\boldsymbol{\mu}_\alpha,\boldsymbol{\Sigma}_\alpha)$ is given by:
\begin{equation}
    \boldsymbol{\Sigma}_\alpha=\left(\boldsymbol{\Sigma}_p \boldsymbol{\Sigma}_g\right)_\alpha^{\boldsymbol{\Sigma}}=\left((1-\alpha) \boldsymbol{\Sigma}_p^{-1}+\alpha \boldsymbol{\Sigma}_g^{-1}\right)^{-1},
\end{equation}
and
\begin{equation}
\begin{aligned}
    \boldsymbol{\mu}_\alpha &=\left(\boldsymbol{\mu}_p \boldsymbol{\mu}_g\right)_\alpha^{\boldsymbol{\mu}} \\
    &=\boldsymbol{\Sigma}_{\alpha} \left((1-\alpha) \boldsymbol{\Sigma}_p^{-1}\boldsymbol{\mu}_p+\alpha \boldsymbol{\Sigma}_g^{-1}\boldsymbol{\mu}_g\right).
\end{aligned}
\end{equation}
Note that $\alpha$ is a parameter that controls the weight of two distributions~\cite{gjsd_2020_entropy} in similarity measurement. In our work, the $\mathcal{N}_p$ and $\mathcal{N}_g$ contribute equally, thus $\alpha$ is set to 0.5. 

Ultimately, for each \textit{gt}, we select $K$ priors which hold the top $K$ GJSD score with this \textit{gt} as the Coarse Positive Samples (CPS) and regard the remaining priors as negative samples, this coarse matching serves as the $\mathcal{M}_{s}$ in Eq.~\ref{c2f_mapping}. The ranking manner works together with the GJSD measurement to construct the Cross-FPN-layer CPS, eliminating the imbalance issue raised by the MaxIoU matching for outlier angles and scales, which will be analyzed in Sec.~\ref{sec.anaysis}.

\subsection{Finer Dynamic Posterior Matching}

Based on Coarse Positive Sample (CPS) candidates, we design a dynamic posterior (prediction) matching rule $\mathcal{M}_d$ to filter out low-quality samples. The $\mathcal{M}_d$ consists of two key components, namely a posterior re-ranking strategy and a Dynamic Gaussian Mixture Model (DGMM) constraint.

We re-rank the sample candidates in the CPS according to their predicted scores. In other words, we further refine the positive samples by their Possibility of becoming True predictions ($PT$)~\cite{ota_2021_cvpr}, which is a linear combination of the predicted classification score and the location score with the \textit{gt}. We define the $PT$ of the $i^{th}$ sample $D_i$ as: 
\begin{equation}
    PT_i = \frac{1}{2}Cls(D_i) + \frac{1}{2}IoU(D_i, gt_i),
\end{equation}
where $Cls$ is the predicted classification confidence and $IoU$ is the rotated IoU between the predicted location and its corresponding \textit{gt} location. We select candidates with $Q$ highest $PT$ as Medium Positive Sample (MPS) candidates. 

Following this, we filter out those samples too far away from the \textit{gts} with a finer instance representation, getting the Finer Positive Samples (FPS).
Different from previous works which utilize the center probability map~\cite{CenterMap-Net_2020_TGRS} or the single-Gaussian~\cite{gaussian3d_2022_pami,gghl_2022_tip} for instance representation, we represent the instance by a finer DGMM. It consists of two components: one is centered on the geometry center and the other is centered on the semantic center of the object. Concretely, for a specific instance $gt_i$, the geometry center $(cx_i,cy_i)$ serves as the mean vector $\boldsymbol{\mu}_{i,1}$ of the first Gaussian, and the semantic center $(sx_i,sy_i)$, which is deduced by averaging the location of the samples in the MPS, serves as the $\boldsymbol{\mu}_{i,2}$. 
That is to say, we parameterize the instance as:
{\small
\begin{equation}
    \setlength{\abovedisplayskip}{\equspace}
    \mathit{DGMM}_i(s|x,y) = \sum_{m=1}^{2}w_{i,m}\sqrt{2\pi|\boldsymbol{\Sigma}_{i,m}|}\mathcal{N}_{i,m}(\boldsymbol{\mu}_{i,m},\boldsymbol{\Sigma}_{i,m}),
\end{equation}
}where $w_{i,m}$ is the weight of each Gaussian with a summation of 1, $\boldsymbol{\Sigma}_{i,m}$ equals to the \textit{gt}'s $\boldsymbol{\Sigma}_{g}$.
Each sample in MPS has a DGMM score $\mathit{DGMM}(s|MPS)$, we set negative masks to samples which have $\mathit{DGMM}(s|MPS) < e^{-g}$ with any \textit{gt}, the $g$ is adjustable.

\section{Experiments}

\subsection{Datasets}
Experiments are done on six datasets, \textit{i.e.},~DOTA-v1.0~\cite{DOTA_2018_CVPR}/v1.5/v2.0~\cite{DOTA2.0_2021_pami}, DIOR-R~\cite{diorr_2022_tgrs}, VisDrone~\cite{visdrone2019_2019_iccvw}, and MS COCO~\cite{COCO_2014_ECCV}.  
In ablation studies and analyses, we choose the large-scale DOTA-v2.0 {\tt train set} for training and {\tt val set} for evaluation, which contains a large number of tiny objects. To compare with other methods, we use {\tt trainval sets} of DOTA-v1.0, DOTA-v1.5, DOTA-v2.0, and DIOR-R for training and their {\tt test sets} for testing, we choose the VisDrone2019, MS COCO {\tt train set}, {\tt val set} for training and~testing.

\setlength{\tabcolsep}{2pt}
\begin{table*}[ht]
\footnotesize
\begin{center}
\resizebox{\linewidth}{!}{
\begin{tabular}{l|c|cccccccccccccccccc|c}  
	\toprule
	Method  & Backbone & Plane & BD & Bridge & GTF & SV & LV & Ship & TC & BC & ST & SBF & RA & Harbor & SP & HC & CC & Air & Heli & mAP  \\
	\midrule
        \textit{multi-stage:}      & & 	&  &  &  &  &  &  &   & & 	&  &  &  &  &  & & & \\
    FR OBB~\cite{Faster-R-CNN_2015_NIPS} & R50 & 71.61 & 47.20 & 39.28 & 58.70 & 35.55 & 48.88 & 51.51 & 78.97 & 58.36 & 58.55 & 36.11 & 51.73 & 43.57 & 55.33 & 57.07 & 3.51 & 52.94 & 2.79 & 47.31 \\
    FR OBB + Dp &  R50 & 71.55 & 49.74 & 40.34 & 60.40 & 40.74 &50.67 &56.58 &79.03 &58.22& 58.24 &34.73 & 51.95 & 44.33 & 55.10 & 53.14 & 7.21 & 59.53 & 6.38 & 48.77 \\
    MR~\cite{Mask-R-CNN_2017_ICCV}  & R50 & 76.20 & 49.91 & 41.61 & 60.00 & 41.08 & 50.77 & 56.24 & 78.01 & 55.85 & 57.48 & 36.62 & 51.67 & 47.39 & 55.79 & 59.06 & 3.64 & 60.26 & 8.95 & 49.47   \\
    HTC*~\cite{HTC_2019_CVPR} & R50 & 77.69 & 47.25 & 41.15 & 60.71 & 41.77 & 52.79 & 58.87 & 78.74 & 55.22 & 58.49 & 38.57 & 52.48 & 49.58 & 56.18 & 54.09 & 4.20 & 66.38 & 11.92 & 50.34 \\
    RT~\cite{RoI-Transformer_2019_CVPR} & R50 &  71.81 & 48.39 & 45.88 & 64.02 & 42.09 & 54.39 & 59.92 & \textcolor{red}{\textbf{82.70}} & \textcolor{blue}{\textbf{63.29}} & 58.71 & 41.04 & 52.82 & 53.32 & 56.18 & 57.94 & 25.71 & 63.72 & 8.70 & 52.81 \\
    Oriented R-CNN~\cite{orientedrcnn_2021_iccv} & R50 & 77.95 & 50.29 & \textcolor{blue}{\textbf{46.73}} & \textcolor{blue}{\textbf{65.24}} & 42.61 & 54.56 & \textcolor{red}{\textbf{60.02}} & 79.08 & 61.69 & 59.42 & 42.26 & \textcolor{red}{\textbf{56.89}} & 51.11 & 56.16 & \textcolor{blue}{\textbf{59.33}} & 25.81 & 60.67 & 9.17 & 53.28\\
    \midrule
      \textit{one-stage:}   & & 	&  &  &  &  &  &  &   & & 	&  &  &  &  &  & & &    \\
    DAL~\cite{dal_2021_aaai} & R50 & 71.23 & 38.36 & 38.60 & 45.24 & 35.42 & 43.75 & 56.04 & 70.84 & 50.87 & 56.63 & 20.28 & 46.53 & 33.49 & 47.29 & 12.15 & 0.81 & 25.77 & 0.00 & 38.52 \\
    SASM~\cite{sasm_2022_aaai} & R50 & 70.30 & 40.62 & 37.01 & 59.03 & 40.21 & 45.46 & 44.60 & 78.58 & 49.34 & 60.73 & 29.89 & 46.57 & 42.95 & 48.31 & 28.13 & 1.82 & \textcolor{blue}{\textbf{76.37}} & 0.74 & 44.53 \\
	RetinaNet-O~\cite{Focal-Loss_2017_ICCV}  & R50 & 70.63 & 47.26 & 39.12 & 55.02 & 38.10 & 40.52 & 47.16 & 77.74 & 56.86 & 52.12 & 37.22 & 51.75 & 44.15 & 53.19 & 51.06 & 6.58 & 64.28 & 7.45 & 46.68    \\
    $\rm{R^3Det}$~w/~KLD~\cite{kld_2021_nips} & R50 & 75.44 & 50.95 & 41.16 & 61.61 & 41.11 & 45.76 & 49.65 & 78.52 & 54.97 & 60.79 & 42.07 & 53.20 & 43.08 & 49.55 & 34.09 & \textcolor{red}{\textbf{36.26}} & 68.65 & 0.06 & 47.26 \\
    FCOS-O~\cite{FCOS_2020_TPAMI} & R50 & 74.84 & 47.53 & 40.83 & 57.41 & 43.89 & 47.72 & 55.66 & 78.61 & 57.86 & 63.00 & 38.02 & 52.38 & 41.91 & 53.24 & 40.22 & 7.15 & 65.51 & 7.42 & 48.51 \\
    Oriented Rep~\cite{orientedrep_2022_cvpr} & R50 & 73.02 & 46.68 & 42.37 & 63.05 & 47.06 & 50.28 & 58.64 & 78.84 & 57.12 & \textcolor{red}{\textbf{66.77}} & 35.21 & 50.76 & 48.77 & 51.62 & 34.23 & 6.17 & 64.66 & 5.87 & 48.95\\
    ATSS-O~\cite{atss_2020_cvpr} & R50 & 77.46 & 49.55 & 42.12 & 62.61 & 45.15 & 48.40 & 51.70 & 78.43 & 59.33 & 62.65 & 39.18 & 52.43 & 42.92 & 53.98 & 42.70 & 5.91 & 67.09 & 10.68 & 49.57 \\
	$\rm{S^2A}$-Net~\cite{s2anet_2021_tgrs}  & R50 & 77.84 & 51.31 & 43.72 & 62.59 & 47.51 & 50.58 & 57.86 & 80.73 & 59.11 & 65.32 & 36.43 & 52.60 & 45.36 & 52.46 & 40.12 & 0.00 & 62.81 & 11.11 & 49.86    \\
    %$\rm{R^3Det + KLD}$ & R50 & - & - & -& - & - & - & - & - & - & - & - & -& - & - & - & - & - & - & 50.90 \\
	\midrule
    \textit{one-stage:}   & & 	&  &  &  &  &  &  &   & & 	&  &  &  &  &  & & &    \\
    DCFL  & R50 & 75.71 & 49.40 & 44.69 & 63.23 & 46.48 & 51.55 & 55.50 & 79.30 & 59.96 & 65.39 & 41.86 & 54.42 & 47.03 & 55.72 & 50.49 & 11.75 & 69.01 & 7.75 & 51.57 \\
    $\rm{S^2A}$-Net w/ DCFL  & R50 & 74.79 & \textcolor{blue}{\textbf{53.25}} & 45.81 & \textcolor{red}{\textbf{65.46}} & 46.49 & 53.23 & 58.10 & \textcolor{blue}{\textbf{81.51}} & 60.13 & \textcolor{blue}{\textbf{66.42}} & 43.24 & 55.09 & 50.52 & 55.58 & 54.53 & 5.23 & 68.73 & 13.06 & 52.84   \\
    DCFL\dag   & R50 & \textcolor{blue}{\textbf{78.30}} & 53.03 & 44.24 & 60.17 & \textcolor{blue}{\textbf{48.56}} & \textcolor{red}{\textbf{55.42}} & 58.66 & 78.29 & 60.89 & 65.93 & \textcolor{blue}{\textbf{43.54}} & 55.82 & \textcolor{blue}{\textbf{53.33}} & \textcolor{red}{\textbf{60.00}} & 54.76 & 30.90 & 74.01 & \textcolor{red}{\textbf{15.60}} & \textcolor{blue}{\textbf{55.08}} \\
    DCFL\dag   & ReR101 & \textcolor{red}{\textbf{79.49}} & \textcolor{red}{\textbf{55.97}} & \textcolor{red}{\textbf{50.15}} & 61.59 & \textcolor{red}{\textbf{49.00}} & \textcolor{blue}{\textbf{55.33}} & \textcolor{blue}{\textbf{59.31}} & 81.18 & \textcolor{red}{\textbf{66.52}} & 60.06 & \textcolor{red}{\textbf{52.87}} & \textcolor{blue}{\textbf{56.71}} & \textcolor{red}{\textbf{57.83}} & \textcolor{blue}{\textbf{58.13}} & \textcolor{red}{\textbf{60.35}} & \textcolor{blue}{\textbf{35.66}} & \textcolor{red}{\textbf{78.65}} & \textcolor{blue}{\textbf{13.03}} & \textcolor{red}{\textbf{57.66}} \\
	\bottomrule
	\end{tabular}}
 \vspace{-2mm}
\caption{Main results on the DOTA-v2.0 OBB Task. We follow the official class abbreviations as the DOTA-v2.0 benchmark\cite{DOTA2.0_2021_pami}. \dag~denotes training for 40 epochs. Note that this paper~\cite{kld_2021_nips} reports 50.90\% mAP for $\rm{R^3Det}$~w/~\rm{KLD} under 20 epochs, the ReR101 backbone is proposed by the ReDet~\cite{redet_2021_cvpr}. The results in \textcolor{red}{\textbf{red}} and \textcolor{blue}{\textbf{blue}} denote the best and second-best performance of each column.}
\vspace{-6mm}
\label{table:dotav2}
\end{center}
\end{table*}
\setlength{\tabcolsep}{1pt}

\begin{table}[t]
    \centering
    \resizebox{\linewidth}{!}{
    \begin{tabular}{cccccc}  
	\toprule
    Method & CFA~\cite{beyond_2021_cvpr} & RetinaNet-O~\cite{Focal-Loss_2017_ICCV} & $\rm{R^3Det}$~\cite{R3Det_2021_AAAI} & Oriented Rep~\cite{orientedrep_2022_cvpr}  & ATSS-O~\cite{atss_2020_cvpr} \\
    mAP  & 69.63 & 69.79 & 70.18 & 71.94 & 72.29 \\
    \midrule
    Method & KLD~\cite{kld_2021_nips} & $\rm{S^2A}$-Net~\cite{s2anet_2021_tgrs} & GGHL~\cite{gghl_2022_tip}(3x) &  DCFL &  DCFL(3x)\\
    mAP & 72.76 & 73.91  & 73.98 & 74.26 & \textbf{75.35} \\
    \bottomrule
    \end{tabular}}
    \vspace{-2mm}
    \caption{Comparison with one-stage detectors on the DOTA-v1.0 OBB Task. All results are based on the MMRotate~\cite{mmrotate_2022_arxiv} with 12 epochs except for GGHL~\cite{gghl_2022_tip}.~3x means training for 36 epochs.}
    \vspace{-2mm}
    \label{exp.dotav1}
\end{table}

\setlength{\tabcolsep}{3pt}
\begin{table}[t]\small
    \centering
    \resizebox{\linewidth}{!}{
    \begin{tabular}{l|c|ccc|c}  
	\toprule
    Method & Backbone  & SV & Ship & ST  & mAP  \\
    \midrule
    RetinaNet-O~\cite{Focal-Loss_2017_ICCV} & R50 & 44.53 & 73.31 & 59.96 & 59.16 \\
    FR OBB~\cite{Mask-R-CNN_2017_ICCV} & R50 & 51.28 & 79.37 & 67.50 & 62.00 \\   
    %MR~\cite{Mask-R-CNN_2017_ICCV} & R50 & 51.31 & 79.75 & 66.07 & 62.67 \\    
    CMR~\cite{Mask-R-CNN_2017_ICCV} & R50 & 51.64 & 79.99 & 67.58 & 63.41 \\
    RT~\cite{RoI-Transformer_2019_CVPR} & R50 & 52.05 & 80.72 & 68.26 & 65.03 \\    ReDet~\cite{redet_2021_cvpr} & ReR50 & 52.38 & 80.92 & 68.64 & 66.86 \\
    \midrule
    DCFL & R50 & 56.72 \textcolor{ForestGreen}{(+12.19)} & 80.87 \textcolor{ForestGreen}{(+7.56)} & 75.65 \textcolor{ForestGreen}{(+15.69)} & 67.37 \textcolor{ForestGreen}{(+8.21)} \\
    DCFL & ReR101 & \textbf{57.31} \textcolor{ForestGreen}{(+12.78)} & \textbf{86.60} \textcolor{ForestGreen}{(+13.29)} & \textbf{76.55} \textcolor{ForestGreen}{(+16.59)} & \textbf{70.24} \textcolor{ForestGreen}{(+11.08)} \\
    \bottomrule
    \end{tabular}}
    \vspace{-2mm}
    \caption{Main results on the DOTA-v1.5 OBB Task. }
    \vspace{-2mm}
    \label{tab.dota15}
\end{table}

\begin{table}[t]
    \centering
    \resizebox{0.95\linewidth}{!}{
    \begin{tabular}{ccccc}  
	\toprule
    Method & RetinaNet-O~\cite{Focal-Loss_2017_ICCV} & FR-OBB~\cite{Faster-R-CNN_2015_NIPS} & RT~\cite{RoI-Transformer_2019_CVPR} & AOPG~\cite{diorr_2022_tgrs}\\
    mAP  & 57.55 & 59.54  & 63.87 & 64.41\\
    \midrule
    Method & GGHL~\cite{gghl_2022_tip} & Oriented Rep~\cite{orientedrep_2022_cvpr}  & DCFL &  DCFL (ReR101)\\
    mAP  & 66.48 & 66.71 & 66.80 & \textbf{71.03} \\
    \bottomrule
    \end{tabular}}
    \vspace{-2mm}
    \caption{Performance comparisons on the DIOR-R dataset.}
    \vspace{-2mm}
    \label{diorr}
\end{table}

\begin{table}[t]\small
    \centering
    \resizebox{0.95\linewidth}{!}{
    \begin{tabular}{l|c|ccc}  
	\toprule
    Method & Backbone & VE & BR & WM \\
    \midrule
    RetinaNet-O~\cite{Focal-Loss_2017_ICCV}  & R50 & 38.0 & 24.0 & 60.2\\
    Oriented Rep~\cite{orientedrep_2022_cvpr}  & R50 & 50.4 & 38.8 & 64.7 \\
    DCFL & R50 & \textbf{50.9} \textcolor{ForestGreen}{(+12.9)} & \textbf{42.1} \textcolor{ForestGreen}{(+18.1)} & \textbf{70.9} \textcolor{ForestGreen}{(+10.7)} \\
    \bottomrule
    \end{tabular}}
    \vspace{-2mm}
    \caption{Detection results of typical tiny objects on the DIOR-R dataset. VE, BR, and WM denote vehicle, bridge, and wind-mill.}
    \vspace{-2mm}
    \label{diorr_class}
\end{table}

\begin{table}[t]\footnotesize
    \centering
    \resizebox{\linewidth}{!}{
    \begin{tabular}{l|cc|cc|cc}  
	\toprule
    Dataset & \multicolumn{2}{c|}{VisDrone} & \multicolumn{2}{c|}{MS COCO} & \multicolumn{2}{c}{DOTA-v2.0 HBB} \\
    \midrule
    Method & RetinaNet~\cite{Focal-Loss_2017_ICCV} & DCFL & RetinaNet & DCFL & FCOS~\cite{rfla_2022_eccv} & DCFL \\
	\midrule
    $\rm{AP}_{0.5}$  & 29.2 & 32.1 & 55.4  & 57.3 &  55.4 & 57.4 \\
	\bottomrule
    \end{tabular}}
    \vspace{-2mm}
    \caption{Results of one-stage object detectors on HBB datasets. }
    \vspace{-4mm}
    \label{visdrone}
\end{table}

\subsection{Implementation Details}
We conduct all the experiments on the computer with a single NVIDIA RTX 3090 GPU, and the batch size is set to 4. Models are built based on MMDetection~\cite{mmdetection_2019_arXiv} and MMRotate~\cite{mmrotate_2022_arxiv} with PyTorch~\cite{PyTorch_2019_NIPS}. The ImageNet~\cite{ImageNet_2015_IJCV} pre-trained models are used as the backbone. The Stochastic Gradient Descent (SGD) optimizer is used for training with a learning rate of 0.005, a momentum of 0.9, and a weight decay of 0.0001. The ResNet-50~\cite{ResNet_2016_CVPR} with FPN~\cite{FPN_2017_CVPR} is the default backbone if not specified. We use Focal loss~\cite{Focal-Loss_2017_ICCV} for classification and IoU loss~\cite{Unitbox_2016_ACMM} for regression. We only use random flipping as data augmentation for all experiments. 

For experiments on the DOTA-v1.0 and DOTA-v2.0, we follow the official settings of the DOTA-v2.0 benchmark~\cite{DOTA2.0_2021_pami}, \textit{i.e.}, we crop images into patches of $\rm{1024 \times 1024}$ with overlaps of 200 and train the model for 12 epochs. For DOTA-v2.0, we reproduce one-stage state-of-the-art methods~\cite{FCOS_2019_ICCV,atss_2020_cvpr,orientedrep_2022_cvpr,R3Det_2021_AAAI,kld_2021_nips,dal_2021_aaai,sasm_2022_aaai,s2anet_2021_tgrs} with the same settings.

For experiments on other datasets, we set the input size to $1024\times1024$ (overlap 200), $800\times800$, $1333\times800$, and $1333\times800$ for DOTA-v1.5, DIOR-R, VisDrone, and COCO respectively. We train the models for 40, 40, 12, and 12 epochs on the DOTA-v1.5, DIOR-R, COCO, and VisDrone as previous works do~\cite{beyond_2021_cvpr,orientedrep_2022_cvpr}. The above settings are fixed unless otherwise specified.

\subsection{Main Results}

\textbf{Results on DOTA series.}
As shown in Tab.~\ref{table:dotav2}, our proposed method achieves the state-of-the-art performance of 57.66\% mAP on the DOTA-v2.0 OBB benchmark under single-scale training and testing. Besides, our model achieves 51.57\% mAP without bells and whistles, surpassing all one-stage object detectors tested. The results on the DOTA-v1.0~\cite{DOTA_2018_CVPR} and DOTA-v1.5 are listed in Tab.~\ref{exp.dotav1}, Tab.~\ref{tab.dota15}. Results also indicate that our DCFL is very effective for detecting tiny oriented objects on the tested datasets, such as small vehicles, ships, and storage tanks, where a boost of about 10 points can be expected compared to the baseline. 

\textbf{Results on DIOR-R.} DIOR-R contains some tiny oriented objects, such as the vehicle, bridge, and windmill. The mAP and class-wise AP of tiny objects are in Tab.~\ref{diorr} and Tab.~\ref{diorr_class}, we also achieve the state-of-the-art performance of 71.03\% mAP and notable improvements on tiny objects.

\textbf{Results on HBB Datasets.} Moreover, we discard the angle to verify the versatility of the DCFL on the generic small object detection datasets VisDrone~\cite{visdrone2019_2019_iccvw}, MS COCO~\cite{COCO_2014_ECCV}, and DOTA-v2.0 HBB~\cite{DOTA2.0_2021_pami}. In Tab.~\ref{visdrone}, our method gets a notable $\rm{AP}_{0.5}$ boost compared to the baseline.

\begin{figure}[t]
\centering
\includegraphics[width=0.9\linewidth]{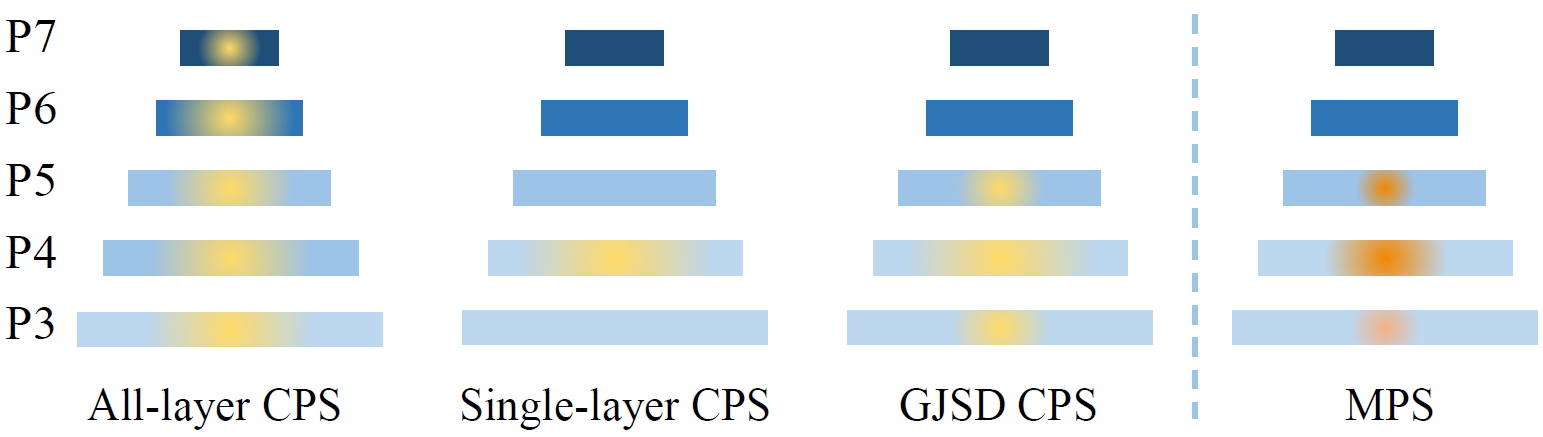}
\caption{Different ways of constructing the CPS. Yellow and orange denote the possible regions of CPS and MPS respectively.}
\vspace{-4mm}
\label{fig:cps}
\end{figure}

\begin{table*}[t]\vspace{-3mm}
% subfloat a - BackBone Architecture
\centering
\subfloat[\textbf{Individual effectiveness.} CPS, MPS, and DGMM denote Coarse, Medium Sample Candidates and Dynamic Gaussian Mixture Model.\label{tab:ablation:individual}]{
\tablestyle{4pt}{1.05}\begin{tabular}{c|ccc|c}  
	\toprule
    Method & CPS & MPS & DGMM & mAP \\
    \midrule
    baseline~\cite{Focal-Loss_2017_ICCV}  &    &  &  &  51.70\\    
    \midrule
      & \checkmark  &  \checkmark  &  & 53.41 \\
     DCFL &  \checkmark  &  &  \checkmark  & 57.20 \\
      &  \checkmark  &  \checkmark  &  \checkmark  & \textbf{59.15} \\
    \bottomrule
    \end{tabular}}\hspace{3mm}
% subfloat b - Multinomial vs Independent Masks
\subfloat[\textbf{Comparsions of different CPS.} The FPN layer number varies for different strategies of getting the CPS.\label{tab:ablation:cps}]{
\tablestyle{4.8pt}{0.95}\begin{tabular}{c|c|c}  
	\toprule
    Strategy & Measurement & mAP  \\
    \midrule
     %base: MaxIoU  & IoU & 51.70 \\
     All-FPN-layer &  Gaussian &  50.12 \\
     Single-FPN-layer &  Gaussian & 56.72 \\
     %Single-FPN-layer &  Center &  \\
     %All-FPN-layer &  Center &  \\
     Cross-FPN-layer & KLD~\cite{kld_2021_nips} &  57.82 \\
     Cross-FPN-layer & GWD~\cite{gwd_2021_icml} &  58.55\\
     Cross-FPN-layer & GJSD &  \textbf{59.15}\\
    \bottomrule
    \end{tabular}}\hspace{3mm}
% subfloat c - RoIAlign (ResNet-50-C4)
\subfloat[\textbf{Effects of detailed designs in the PCB.}\label{tab:ablation:pcb} DP denotes the dynamic prior. Guiding denotes \textit{reg} guides \textit{cls}.]{
\tablestyle{2.2pt}{0.95}\begin{tabular}{ccc|c}  
	\toprule
    DCN & Dilated Conv & DP & mAP \\
    \midrule
     & & & 58.07 \\
    \checkmark & & & 58.41\\
    \checkmark & \checkmark & & 58.65\\
    \textit{Separate} & \checkmark & \checkmark & 58.71 \\
    \textit{Guiding} & \checkmark & \checkmark & \textbf{59.15} \\
    \bottomrule
    \end{tabular}}\\
% subfloat d - Effects of parameters 
\subfloat[\textbf{Effects of parameters $K$ and $Q$.}\label{tab:ablation:kq}]{
\tablestyle{4.5pt}{0.55}\begin{tabular}{c|cccc|cccc}  
	\toprule
    $K$ & \multicolumn{4}{|c|}{24} & \multicolumn{4}{c}{20}  \\
    \midrule
    $Q$  &  20 & 16 & 12 & 8 &  16 & 12 & 10 & 8 \\
    \midrule
    mAP  &  58.31 & 58.11 & 58.95 & 59.06 &  58.66 & 58.71 & 58.92 & 58.28 \\
    \midrule
    \midrule
    $K$ & \multicolumn{4}{|c|}{16} & \multicolumn{4}{c}{12}   \\
    \midrule
    $Q$  &  12 & 10 & 8 & 6 &  10 & 8 & 6 & 4 \\
    \midrule
    mAP  &  \textbf{59.15} & 58.57 & 58.97 & 57.84 &  58.79 & 58.25 & 57.01 & 57.37\\
    \bottomrule
    \end{tabular}}\hspace{3mm}
% subfloat e - mask representation
\subfloat[\textbf{Effects of parameter $g$}.\label{tab:ablation:g}]{
\tablestyle{10pt}{1.38}\begin{tabular}{c|cc}  
    	\toprule    
        $g$ & 1.2 & 1.0 \\
        \midrule
        mAP & 57.91 &  58.20  \\
        \midrule
        \midrule
        $g$ & 0.8 & 0.4\\
        \midrule
        mAP & \textbf{59.15} & 58.95 \\
        \bottomrule
        \end{tabular}}
% main caption
\caption{\textbf{Ablations}. We train on DOTA-v2.0 \texttt{train set}, test on \texttt{val set}, and report mAP under IoU threshold 0.5.}
\label{tab:ablations}\vspace{-3mm}
\end{table*}

\subsection{Ablation Study}

\textbf{Effects of Individual Strategy.} We check the effectiveness of each proposed strategy in the proposed method. In all ablation experiments, we employ one prior for each feature point for fair comparisons. As seen in Tab.~\ref{tab:ablation:individual}, the baseline detector RetinaNet-OBB yields a result of 51.70\% mAP. When we gradually apply the posterior re-ranked MPS and DGMM into the detector based on the CPS, the mAP improves progressively, verifying each design's effectiveness. Note that the CPS cannot be independently used since the samples in it are too coarse to serve as the final positive samples. Nevertheless, we compare some different ways of constructing the CPS to verify its superiority.   

\textbf{Comparisons of Different CPS.} The design of the CPS matters in the training pipeline. We show several paradigms of designing the CPS as shown in Fig.~\ref{tab:ablation:cps}, including limiting the CPS for a specific \textit{gt} within a single layer, releasing all FPN layers as the CPS, like Objectbox~\cite{objectbox_2022_eccv}. We compare their performance in Tab.~\ref{tab:ablation:cps}. For fair comparisons, the number of samples in CPS is fixed at 16, and all other components are kept the same. For the Single-FPN-layer way, we group \textit{gt} onto different layers according to the scale division strategy in FCOS, then assign labels within each layer. For the All-FPN-layer way, we do not group \textit{gt} onto different layers, instead, we discard the prior scale information and directly measure the distance between Gaussian \textit{gt} and prior points. The results are shown in Tab.~\ref{tab:ablation:cps}, 
we can observe that neither of the above two ways will yield the best performance. By contrast, the distribution distances (KLD, GWD, GJSD) are able to construct the Cross-FPN-layer CPS, where the candidates are extended to adjacent layers besides the main layer. We can also see the GJSD gets the best performance of 59.15\% mAP, mainly resulting from its property of scale-invariance~\cite{kld_2021_nips,gjsd_2020_entropy}, symmetry~\cite{gjsd_2020_entropy}, and ability to measure non-overlapping boxes~\cite{gjsd_2020_entropy} compared to other counterparts.

\textbf{Fixed Prior and Dynamic Prior.} We conduct a finer group of ablation studies to verify the necessity of introducing the dynamic prior. As shown in Tab.~\ref{tab:ablation:pcb}, if we disable the dynamic prior by fixing the location of samples, a slight performance drop will be introduced. Hence, the prior should be adjusted accordingly when leveraging the dynamic sampling strategy to better capture the shape of objects.

\textbf{Detailed Design in PCB.} For the PCB, it is made up of a dilated convolution and a guiding DCN, we slightly enlarge the receptive field with a dilation rate of 3. After that, we take advantage of the DCN to generate dynamic priors in a guiding manner. As shown in Tab.~\ref{tab:ablation:pcb}, we can observe that the DCN can bring an improvement of 0.34 mAP points and the dilated convolution can slightly enhance the mAP. We find that the application of the DCN~\cite{DCN_2017_CVPR} to the single regression branch will slightly deteriorate the accuracy (noted by \textit{Separate} in Tab.~\ref{tab:ablation:pcb}), which may cause mismatch issues between the two branches. Thus we utilize the offsets from the regression head to guide the offsets classification head for better alignment (noted by \textit{Guiding}).

\textbf{Effects of Parameters.} The introduced three parameters are robust in a certain range. From Tab.~\ref{tab:ablation:kq}, we can see that a combination of $K=16$ and $Q=12$ gets the best performance. In Tab.~\ref{tab:ablation:g}, we verify the threshold $e^{-g}$ in the DGMM, we empirically set $w_{i,1}$ to 0.7, then a threshold of $g=0.8$ yields the highest mAP. Although making the CPS/MPS/FPS coarser and stricter will weaken the performance, the mAP only waves marginally. In other words, the coarse-to-fine assignment manner somewhat warrants the parameter selection's robustness since multiple parameters can attenuate the effects of an under-tuned one.

\begin{figure*}[t]
\centering
\includegraphics[width=0.95\linewidth]{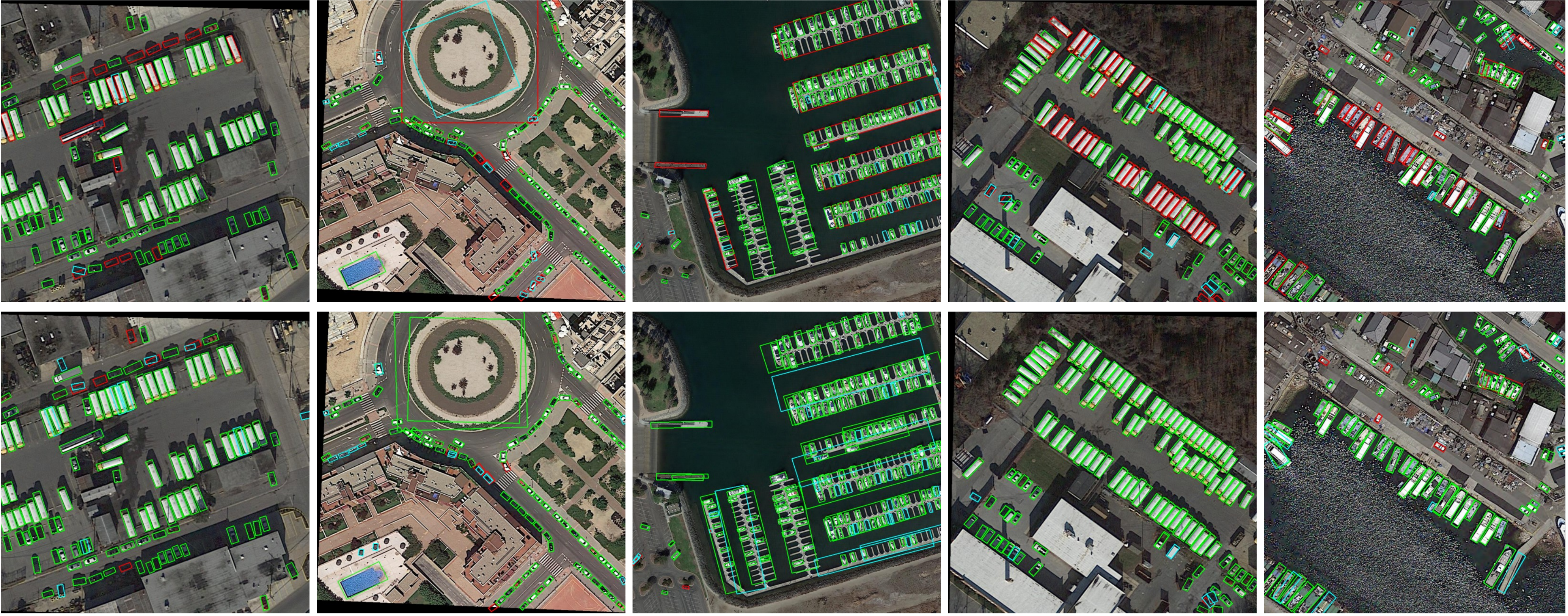}
\vspace{-2mm}
\caption{Visualization analysis of the predicted results. The first row is the result of the RetinaNet-OBB while the second row is the result of the DCFL. TP, FN, and FP predictions are marked in green, red, and blue respectively. }
\label{fig:vis}
\vspace{-3mm}
\end{figure*}

\begin{figure}[]
\centering
\includegraphics[width=0.95\linewidth]{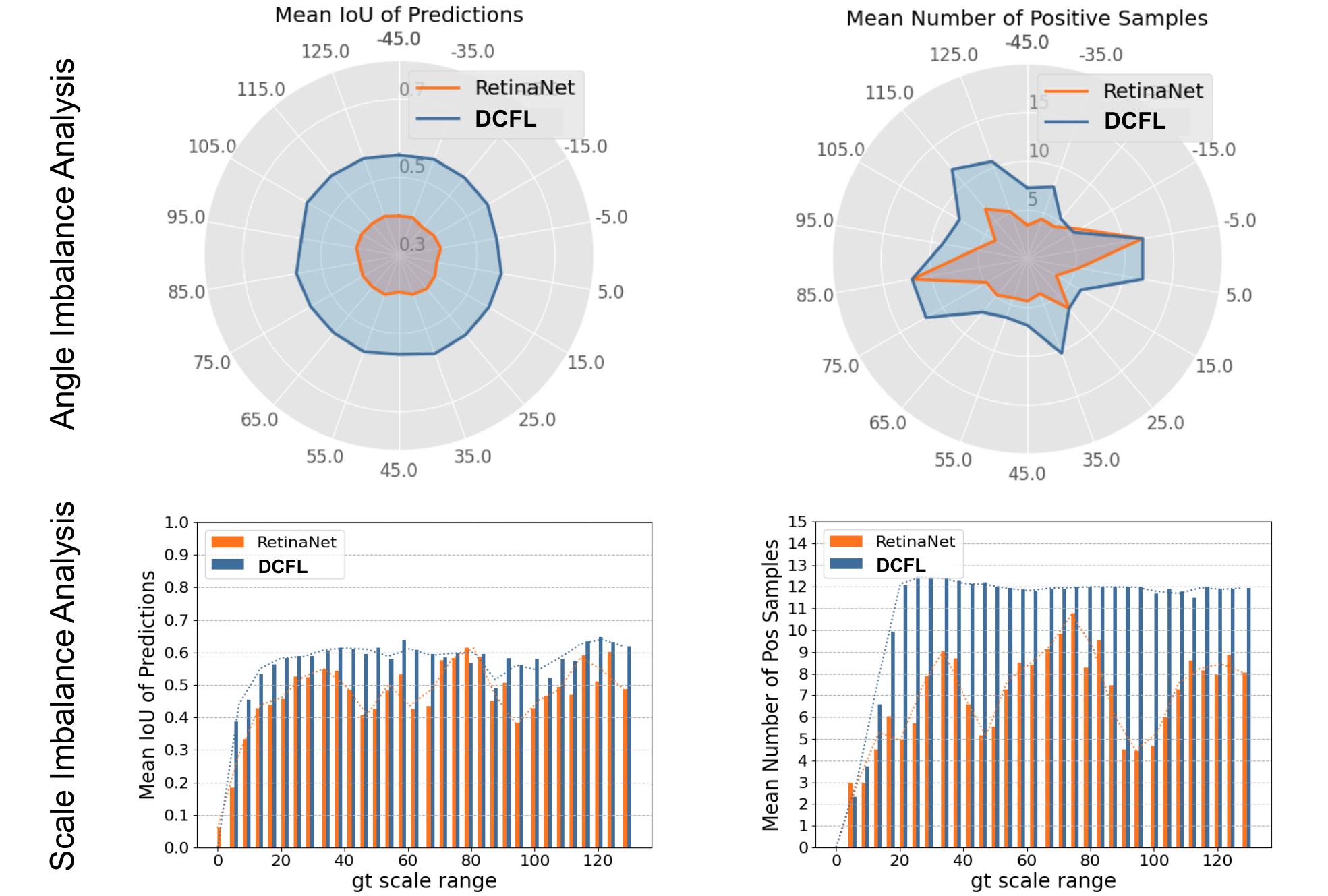}
\vspace{-2mm}
\caption{Statistical analysis of imbalance issues. The first and second columns show quality and quantity imbalance  respectively.}
\label{fig:quality_imbalance}
\vspace{-3mm}
\end{figure}

\begin{figure}[]
\centering
\includegraphics[width=0.9\linewidth]{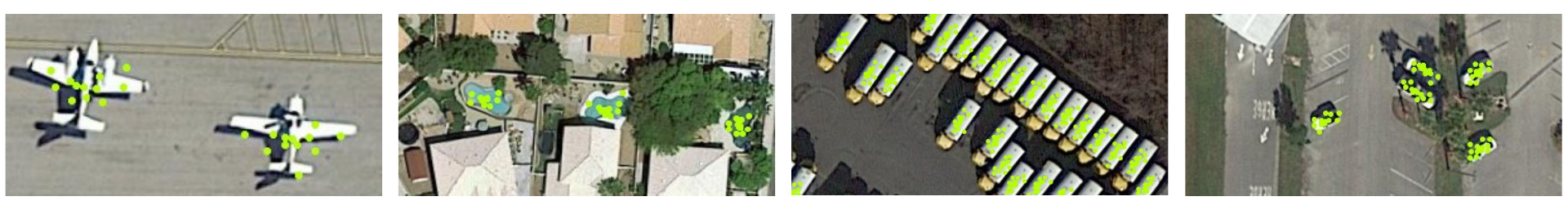}
 \vspace{-4mm}
\caption{Visualization of sampled dynamic priors.}
\label{fig:vis_samples}
\vspace{-6mm}
\end{figure}

\section{Analysis}
\label{sec.anaysis}
For a clearer dissection of why the proposed scheme works, we perform more meticulous analyses as follows.

\textbf{Reconciliation of imbalance problems.} To delve into the imbalance issue, we calculate the mean predicted IoU and the mean positive sample number of \textit{gt} holding different angles and different scales (absolute size). Results are shown in Fig.~\ref{fig:quality_imbalance}, which are from the models' last training epoch. Here we summarize two kinds of imbalance issues (quantity and quality imbalance) for RetinaNet: (1) The positive sample number assigned to each instance changes periodically \textit{w.r.t.}~its angle and scale, whereas objects with shapes (scale, angle) different from predefined priors will hold much fewer positive samples. (2) The predicted IoU changes periodically \textit{w.r.t.}~\textit{gt}'s scale while remaining invariant \textit{w.r.t.}~\textit{gt}'s angle. By contrast, DCFL remarkably reconciles the imbalance: (1) more positive samples are compensated to previously outlier angles and scales. (2) the samples' quality (predicted IoU) can also be improved and balanced across all angles and scales. The above results are the desired behavior of dynamic coarse-to-fine learning. %The reconciliation of imbalance issues serves as strong evidence for the effectiveness of DCFL.

\begin{table}[t]%\footnotesize
    \centering
    \resizebox{\linewidth}{!}{
    \begin{tabular}{c|ccccc}    
	\toprule
        Method &  $\rm{R^3Det}$~\cite{R3Det_2021_AAAI} & $\rm{S^2A}$-Net~\cite{s2anet_2021_tgrs} & GA-RetinaNet~\cite{guidedanchoring_2019_cvpr} & RetinaNet~\cite{Focal-Loss_2017_ICCV} & DCFL  \\
        \midrule
        Params, GFLOPs & 42.0M, 337.3 & 38.6M, 197.9 & 37.4M, 206.9 & 36.5M, 217.3 & \textbf{36.1M, 157.8} \\
	\bottomrule
    \end{tabular}}
    \vspace{-4mm}
    \caption{Comparison of \textit{params}, GFLOPs with $1024 \times 1024$ input. }
    \vspace{-6mm}
    \label{tab.params}
\end{table}

\textbf{Visualization.} We visualize the predicted results and positive samples in Fig.~\ref{fig:vis} and Fig.~\ref{fig:vis_samples}. We can see that the DCFL remarkably eliminates the False Negative and False Positive predictions, especially for the extreme-shaped oriented tiny objects. Fig.~\ref{fig:vis_samples} shows that the proposed strategy is able to dynamically generate and sample priors that better fit the instance's main body, verifying the claims of dynamic modeling and mismatch alleviation in this work.

\textbf{Speed.} We test the inference speed on DOTA-v2.0 {\tt val set} with a single RTX3090 GPU, the FPS of the $\rm{R^3Det}$, $\rm{S^2A}$-Net, RetinaNet, and DCFL is 16.2, 18.9, 20.8, and 20.9. It indicates that our method is of high efficiency. Moreover, we provide the parameters and GLOPs in Tab.~\ref{tab.params}, where we can see that the DCFL is lighter.

\vspace{-2mm}
\section{Conclusion}
In this paper, we propose a novel DCFL scheme for detecting oriented tiny objects. We identify that the mismatched feature prior and unbalanced positive samples are two obstacles hampering the label assignment for oriented tiny objects. To address these, we propose a dynamic prior to alleviate the mismatch issue and a coarse-to-fine assigner to mitigate the imbalance issue, where the prior, label assignment, and \textit{gt} representation are all reformulated in a dynamic manner. Extensive experiments and analyses show the convincing improvements brought by the DCFL. 

\section*{Acknowledgement}
We thank the reviewers for their comments. This work was supported in parts by NSFC (62271355, 62271354) and the Fundamental Research Funds for the Central Universities (2042022kf1010).

%%%%%%%%% REFERENCES
{\small
\bibliographystyle{ieee_fullname}
\bibliography{bibs/chang, bibs/jinwang}
}

\end{document}